\documentclass[journal]{IEEEtran}
\usepackage{cite}
\usepackage{xcolor}
\usepackage{graphicx}
\usepackage{multirow}
\usepackage{hyperref}
\usepackage[switch]{lineno}
\usepackage{makecell}

\begin{document}
\bibliographystyle{IEEEtran}
\title{Deep Learning-based Development of Personalized Human Head Model with Non-Uniform Conductivity for Brain Stimulation}
    
\author{Essam~A.~Rashed,~\IEEEmembership{Senior Member,~IEEE,}
            Jose~Gomez-Tames,~\IEEEmembership{Member,~IEEE}
            and~Akimasa~Hirata,~\IEEEmembership{Fellow,~IEEE}
\thanks{This work was supported in part by the Ministry of Internal Affairs and Communications, Japan.
E. A. Rashed, J. Gomez-Tames, and A. Hirata are with the Department of Electrical and Mechanical Engineering, Nagoya Institute of Technology, Nagoya 466-8555, Japan. E. A. Rashed is also with the Department of Mathematics, Faculty of Science, Suez Canal University, Ismailia 41522, Egypt. e-mail: \href{mailto:essam.rashed@nitech.ac.jp}{essam.rashed@nitech.ac.jp}}}
\maketitle


\begin{abstract}
Electromagnetic stimulation of the human brain is a key tool for the neurophysiological characterization and diagnosis of several neurological disorders. Transcranial magnetic stimulation (TMS) is one procedure that is commonly used clinically. However, personalized TMS requires a pipeline for accurate head model generation to provide target-specific stimulation. This process includes intensive segmentation of several head tissues based on magnetic resonance imaging (MRI), which has significant potential for segmentation error, especially for low-contrast tissues. Additionally, a uniform electrical conductivity is assigned to each tissue in the model, which is an unrealistic assumption based on conventional volume conductor modeling. This paper proposes a novel approach to the automatic estimation of electric conductivity in the human head for volume conductor models without anatomical segmentation. A convolutional neural network is designed to estimate personalized electrical conductivity values based on anatomical information obtained from T1- and T2-weighted MRI scans. This approach can avoid the time-consuming process of tissue segmentation and maximize the advantages of position-dependent conductivity assignment based on water content values estimated from MRI intensity values. The computational results of the proposed approach provide similar but smoother electric field results for the brain when compared to conventional approaches.

\end{abstract}


\IEEEpeerreviewmaketitle


\section{Introduction}

\begin{figure*}
\centering
\includegraphics[width=\textwidth]{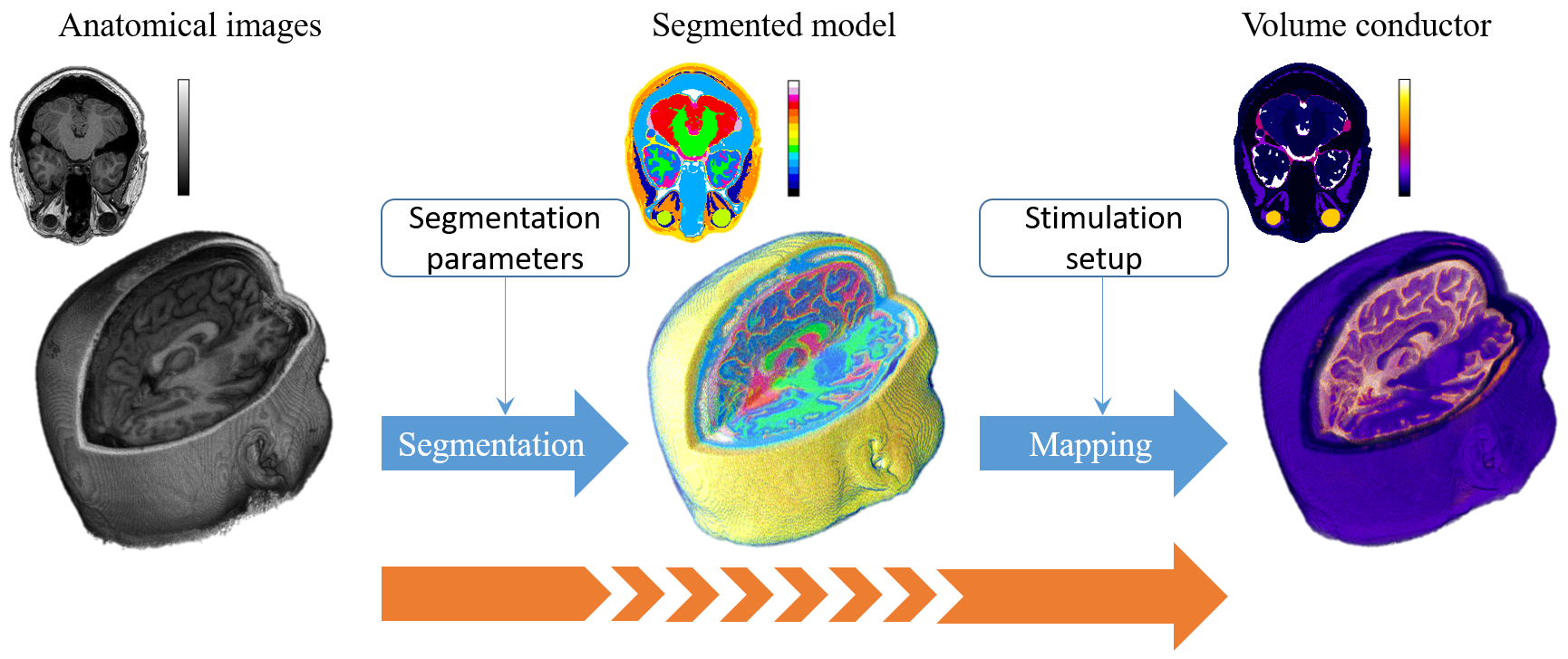}
\caption{Standard pipeline for the generation of volume conductor models with uniform conductivity based on anatomical images for dosimetry computations. Orange arrows indicate the contributions of this study where segmentation is not required and non-uniform conductivity map is estimated.}
\label{fig01}
\end{figure*}

In electromagnetic dosimetry applications, the use of computational models that imitate human anatomy is an essential process~\cite{Kainz2019TRPMS1}. Such models are used to simulate biological tissues as volume conductors for various electromagnetic characterization and neuromodulation applications, as well as human safety studies. A current major trend in healthcare services is precision medicine, where medical decisions, treatments, or diagnoses are customized to fit the characteristics and conditions of individual subjects. Recent developments in medical imaging have led to more personalized data for precision medicine applications related to brain disorders~\cite{Inse2015science, Comaniciu2016MIA}. For electromagnetic stimulation of the brain, the formulation of personalized head models is an emerging trend with the goal of avoiding the significant inconsistency caused by inter- and intra-subject variability~\cite{Rashed2019PULSE}. The current standard pipeline for human head modeling begins with the acquisition of anatomical images (magnetic resonance imaging (MRI) or computed tomography), followed by intensive segmentation of different tissue compositions. Therefore, uniform tissue conductivity is assigned to each annotated tissue (Fig.~\ref{fig01}). In most cases, isotropic conductivity is assumed to be valid for almost all structures. This approach has several limitations that could lead to incorrect estimation of \emph{in situ} electric fields, reducing the accuracy of stimulation planning. First, the human head is known to be composed of different biological tissues with a wide range of conductivity values~\cite{McCann2019BT}. Some of these tissues, especially non-brain tissues, exhibit low contrast or only appear within limited regions in anatomical images. Therefore, segmentation of all tissues is a challenging task that is difficult to perform accurately. Second, uniform conductivity is an unrealistic assumption because even within the same tissue, conductivity values can vary based on different parameters, such as water content~\cite{Michel2017MRM}, sodium concentration~\cite{Liao2019MRM}, and anatomical structure. For example, skin conductivity is known to vary significantly from the surface layer to deeper layers~\cite{Wake2016PMB}. The use of uniform conductivity values may provide a reasonable approximation and is widely used in dosimetry studies, but it is still unrealistic, especially within tissue border regions~\cite{Gurler2017MRM}. Third, conductivity values are typically derived from measurements presented in literature (e.g., \cite{Gabriel1996PMB}), making it difficult to personalize such values based on measurement conditions, measurement methods, temperature, subject age, etc.~\cite{Wang2006TEMC,Peyman2007PMB}.

In our previous work, we developed an efficient deep learning architecture called ForkNet for the fast and efficient segmentation of several human head tissues~\cite{Rashed2019neuroimage, Rashed2019ICIP}. ForkNet can provide high-quality segmentation results that fit the pipeline presented in Fig.~\ref{fig01}. However, limitations appear based on the use of uniform conductivity assumptions for each tissue, which are also used in similar frameworks. To avoid potential errors caused by segmentation faults, several methods have been proposed to estimate electrical conductivity based on anatomical images (mainly MRI)~\cite{Ropella2017MRM, Elsaid2017TMI, Hampe2018MRM, Serralles2019TBE1}. The water content calculated from T1-weighed MRI scans is modeled by a monotonic function to estimate the conductivity of major brain tissues~\cite{Michel2017MRM}. However, such methods utilize data that is strictly limited to brain tissues. Magnetic resonance electrical impedance tomography (MREIT) was presented as a useful approach for estimating brain conductivity~\cite{Chauhan2018TMI, Kwon2016TBE1, Kwon2014PMB}, but it is also strictly limited to brain tissues. A recent review of the methods used to estimate electrical conductivity based on MRI was presented in~\cite{Liu2017TBE}.

Deep learning has become a standard machine learning technique for several data mapping and labeling problems~\cite{Lecun2015Nature}. Specifically, convolutional neural networks (CNNs) are now the leading tool for image processing and recognition. This success motivated us to investigate deep learning approaches to estimating electric conductivity based on MRI for electromagnetic dosimetry computations and other applications. 

In this paper, we propose a method for the automatic generation of volume conductor models without the segmentation of complicated head tissues. To the best of our knowledge, this paper presents the first method for the automatic estimation of non-uniform electric conductivity in human head models based on deep learning. The proposed method maps T1- and T2-weighted MRI scans to potential electrical conductivity values.


\section{Uniform versus non-uniform conductivity}

In non-invasive electrostimulation, the simulated distributions of electric fields in the brain can be used to estimate the effects of stimulation sessions \cite{Balslev2007JNM}. Simulation studies based on volume conductor models with finite-element or finite-difference methods are commonly used to solve Maxwell's equation. Therefore, accurate electrical conductivity predictions for different tissues in the head are required to derive precise maps of electric field distributions in the brain. However, the estimation of accurate conductivity values is difficult because it requires reliable measurements from live homogeneous tissues. It has become a common practice in this field to use uniform conductivity values within consistent segmented tissues.

\begin{figure}
\centering
\includegraphics[width=0.5\textwidth]{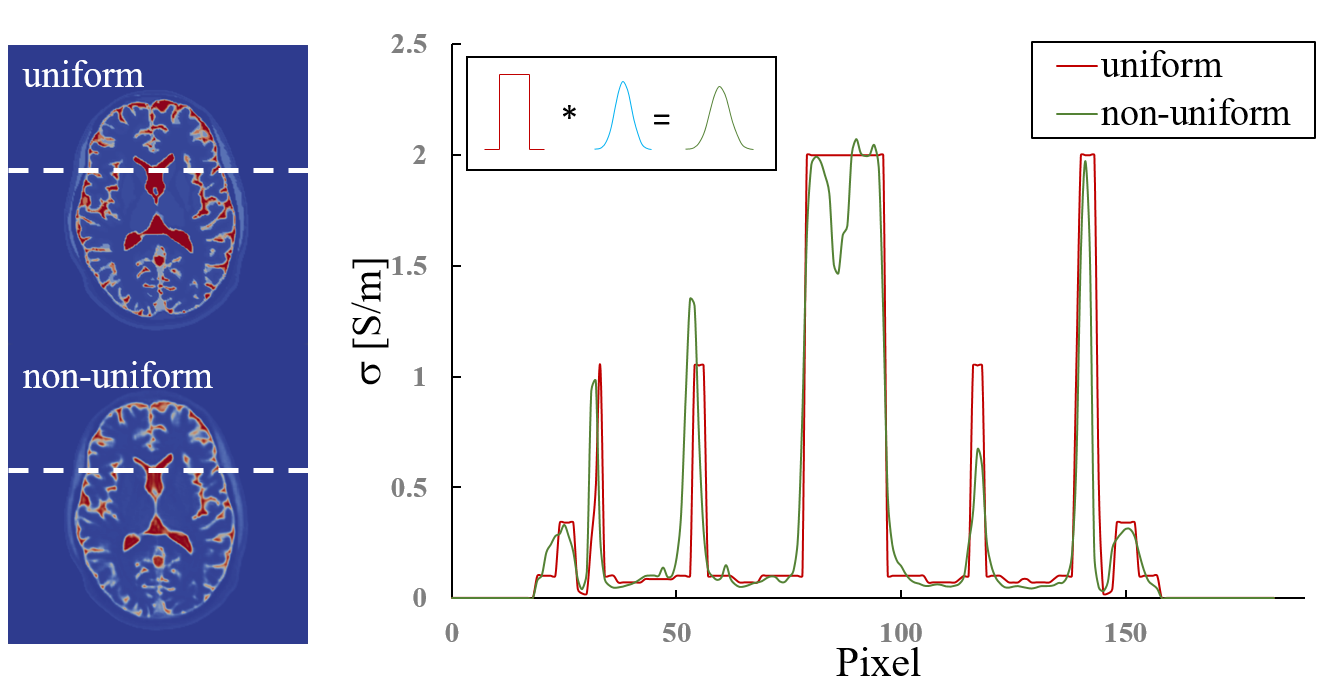}
\caption{Example of the differences between uniform and non-uniform electrical conductivity estimation. Sample slices are presented on the left and the profiles of the white dashed lines are plotted on the right.}
\label{fig02}
\end{figure}

\begin{figure*}
\centering
\includegraphics[width=\textwidth]{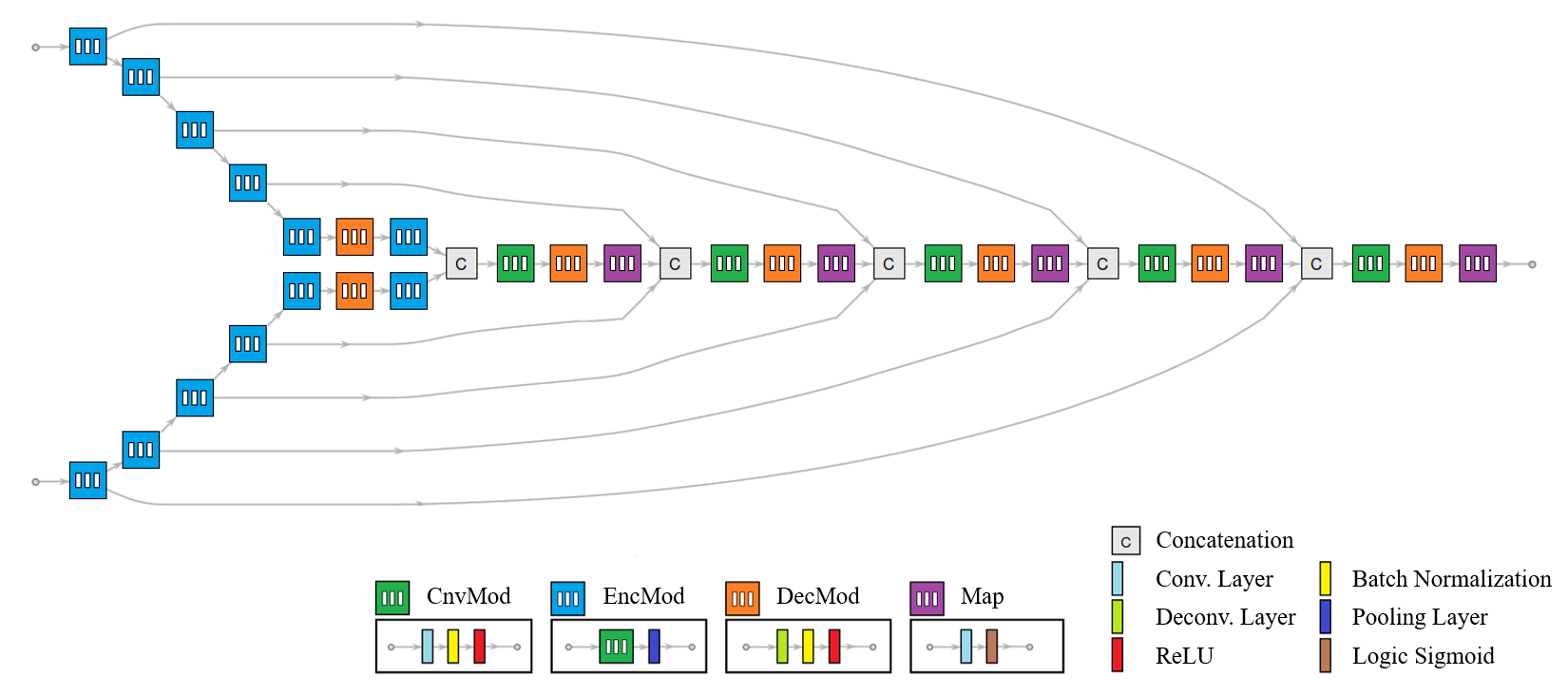}
\caption{Architecture of the proposed CondNet with layer identification keys. This architecture consists of two inputs ($U = 2$) and a single output ($V = 1$) with a and depth of $I=6$. Detailed feature variables for each layer are listed in Table~\ref{tab02}}.
\label{fig03}
\end{figure*}

Although uniform conductivity for each tissue is widely used, it has certain limitations. To demonstrate the difference between uniform and non-uniform conductivity, we will present a single slice of two electrical conductivity maps, where one was generated using uniform conductivity (computed using the method in~\cite{Laakso2015BS}) and the other was generated using non-uniform conductivity (computed using the method presented in this paper). The differences in conductivity values are summarized by the profile plots in Fig.~\ref{fig02}. Uniform conductivity produces sharp edges that indicate sudden changes in conductivity values at borders, whereas non-uniform conductivity produces much smoother results. This demonstration makes it clear that non-uniform profiles can be interpreted as Gaussian convolutions of uniform profiles. From a computational complexity perspective, using uniform conductivity seems to be straightforward because it simply requires segmentation of anatomical images, after which each tissue can be assigned a single conductivity value. However, it is expected that segmentation errors could have a significant influence on electric field maps, especially at the border regions between tissues with similar anatomical gray values and different conductivity measurements (e.g, muscle and gray matter (GM) or cerebrospinal fluid (CSF) and air). In such cases, segmentation errors are likely to generate relatively large errors in conductivity values. Furthermore, sudden changes in conductivity values at border regions can produce staircase artifacts in electric field maps. Averaging techniques and the inclusion of specific percentile values are common post-processing techniques for avoiding such artifacts~\cite{Hirata2010PMB, GomezTames2018TEMC}. In contrast, non-uniform conductivity represents a more personalized approach because the conductivity at each discrete point (voxel) is assumed to be independent. The estimation of non-uniform conductivity using MRI was first proposed based on the estimation of the radio frequency penetration of MRI~\cite{Haacke1991PMB}. Recently, the electrical conductivity of brain tissues has been estimated using B1 maps from 7T MRI~\cite{Wang2019MRI}. Almost all approaches to non-uniform conductivity estimation are based on highly sophisticated imaging modalities that place additional burdens on patients. In this study, we aimed to develop a simple and effective method to estimate electrical conductivity using standard imaging modalities, such as T1- and T2-weighted MRI.


\section{Materials and methods}


\subsection{Data and preprocessing}

A set of freely available T1- and T2-weighted MRI scans containing 256$^3$ voxels with a uniform voxel size of 1.0~mm$^3$ was utilized in this study\footnote{http://hdl.handle.net/1926/1687}. The semi-automatic method presented in~\cite{Laakso2015BS} was used to segment MRI scans from 18 subjects into different tissues, such as skin, muscle, fat, bone (cortical), bone (cancellous), dura, blood, CSF, GM, white matter (WM), cerebellum, vitreous humor, and mucous tissue. The dataset was split arbitrarily into 10 subjects for training and 8 subjects for testing. The number of subjects required for training was computed in our previous study and we determined that 10 to 15 subjects are sufficient for stable training~\cite{Rashed2019neuroimage}. A bias correction is implemented for both T1 and T2 MRI~\cite{Laakso2015BS}. Segmented models are assigned to isotropic uniform tissue conductivity values using a fourth-order Cole-Cole model with a frequency of 10 kHz for transcranial magnetic stimulation (TMS) applications, as reported in~\cite{Gabriel1996PMB}. The uniform conductivity values are listed in Table~\ref{tab01}~(A). MRI scans are normalized individually such that they have zero mean and unit variance, followed by scaling in the range of $[0, 1]$. The corresponding uniform volume conductor is scaled to range of $[0, 1-\tau]$, where $\tau$ is a small value parameter (here, $\tau = 0.1$).

\begin{table}
\centering
\footnotesize
\caption{Human tissue conductivity values [S/m] for (A) the Cole-Cole model and (B) typical values from computational studies \cite{Aonuma2018neuroimage1}.}
\setlength{\tabcolsep}{3pt}
\begin{tabular}{ l c c  |l  c c  }
\hline
\multirow{2}{*}{\bf Tissue}	& \multicolumn{2}{c|}{ Conductivity $\sigma_n$}   & \multirow{2}{*}{\bf Tissue}		& \multicolumn{2}{c}{ Conductivity $\sigma_n$}	\\
\cline{2-3} \cline {5-6}
& A & B & & A & B\\
\hline
Blood            & 0.700 & 0.700 & GM                   & 0.100 & 0.276 \\
Bone (canc.) & 0.080 & 0.025 & Mucous tissue & 0.070 & 0.070 \\
Bone (cort.)  & 0.020 & 0.007 & Muscle             & 0.340 & 0.400 \\ 
Cerebellum  & 0.130 & 0.276 & Skin                  & 0.100 & 0.456 \\
CSF             & 2.000 & 1.654 & V.  Humor         & 1.500 & 1.500 \\
Dura            & 0.500 & 0.500 & WM                   & 0.070 & 0.126 \\
Fat              & 0.040 & 0.040 & 			        &           &          \\
\hline
\end{tabular}
\label{tab01}
\end{table}


\subsection{Network architecture design}

The proposed network architecture, which is called a conductivity network (CondNet), is a multitask end-to-end mapping architecture that can generally connect $U$ anatomical images to $V$ volume conductors. A simple example is presented in Fig.~\ref{fig03} with $U = 2$, $V = 1$, and network depth $I = 6$. The feature size for each layer is detailed in Table~\ref{tab02}. The design of CondNet is based on individual encoder tracks from several convolutional operations aimed at extracting features from anatomical images. The decoder tracks assigned to volume conductors are generated through a series of convolutional and deconvolutional operations with feedback features (skip connections) from all encoder tracks. As detailed in Table~\ref{tab02}, the convolutional kernels ($R_{u,i}$, $S_{v,i}$, and $T_{v,i}$) can be customized to fit feature selection and anatomical textures in different network layers. Considering $M^1, M^2,\dots,M^U$ are the network normalized input volumes, each of which has $K$ slices, and the output of the CondNet is computed as follows:
\begin{equation}
\{L_{k}^{1},L_{k}^{2},\dots,L_{k}^{V}\}=\textnormal{CondNet}(M_k^1,M_k^2,\dots,M_k^U), \forall k,
\label{eq1}
\end{equation}
where $M_k^u$ is an anatomical slice from input image $u$ and $L_k^v$ is the corresponding slice from the normalized volume conductor $L^v$. Considering different slicing directions, the average normalized volume conductor is computed as the mean value obtained from different orientations as follows:

\begin{equation}
L^v_{*}= \frac{1}{3}(L^v_{a}+L^v_{s}+L^v_{c}), \forall v,
\label{eq2}
\end{equation}
where $L_{a}$, $L_{s}$, and $L_{c}$ are normalized volume conductors computed in the axial, sagittal, and coronal directions, respectively. Finally, a standard volume conductor is computed as follows:

\begin{equation}
C^v=\frac{\max_n({\sigma_n^v})}{1-\tau} L^v_{*}, \forall n, v,
\label{eq3}
\end{equation}
where $\sigma_n^v$ is the electrical conductivity value assigned to tissue $n$ of volume conductor $v$.

\begin{table*}
\centering
\footnotesize
\caption{Detailed architecture of CondNet (shown in Fig.~\ref{fig03}) with $U$ inputs, $V$ outputs, and a depth of $I$. The convolution kernels $R$, $S$, and $T$ can be customized individually.}
\begin{tabular}{|l|lll|c|}
\hline 
{\bf Module} & {\bf Layer} & {\bf Output size} & {\bf Kernel} &   {\bf Label}\\
\hline \hline
Input$_u$ & & \multirow{2}{*}{$[2^8]^2$} &&\\
$u:1 \rightarrow U$ & &  &&\\
\hline
\hline
EncMod$_{u,i}$ & Convolution & $2^{(i+1)} \times [2^{(8-i)}]^2 $ & $2^{(i+1)} \times [R_{u,i}]^2 $ &  \multirow{3}{*}{\includegraphics[width=.5cm]{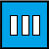}}\\
$u:1 \rightarrow U$& BN \& ReLU & $2^{(i+1)} \times [2^{(8-i)}]^2 $ &   & \\
$i:1 \rightarrow I$& Pooling (Max) & $2^{(i+1)} \times[2^{(7-i)}]^2 $ && \\
\hline
\hline
\multirow{2}{*}{Hub} & \multirow{2}{*}{Concatenation} & \multirow{2}{*}{$U \times 2^{I} \times[2^{(9-I)} ]^2 $} & &\multirow{2}{*}{\includegraphics[width=.4cm]{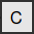}}\\
 &  &   &&\\
\hline
\hline
DecMod$_{v,i}$ & Deconvolution & $2^{(i+1)} \times [2^{(9-i)}]^2$ & $2^{(9-i)}$ $\times$2$\times$2 & \multirow{3}{*}{\includegraphics[width=.5cm]{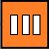}} \\
$v:1 \rightarrow V$ & BN \& ReLU & $2^{(i+1)} \times [2^{(9-i)}]^2$ &   & \\
$i:(I-1) \rightarrow 1$ &  &   &     &\\
\hline
\hline
CnvMod$_{v,i}$ & Convolution & $2^{(i+2)} \times [2^{(8-i)}]^2 $ & $2^{(8-i)} \times [ S_{v,i}]^2$ & \multirow{3}{*}{\includegraphics[width=.5cm]{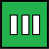}}\\
$v:1 \rightarrow V$ & BN \& ReLU &$2^{(i+2)} \times [2^{(8-i)}]^2 $ &   & \\
$i:(I-1) \rightarrow 1$ & & & &  \\
\hline
\hline
Map$_{v,i}$ & Convolution & $2^i \times [2^{(9-i)}]^2$ & $2^{(9-i)} \times [ T_{v,i}]^2$&\multirow{4}{*}{\includegraphics[width=.5cm]{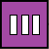}}\\
$v=1 \rightarrow V$  & Sigmoid (Log) &$ \left\{\begin{array}{ll}
2^i \times [2^{(9-i)}]^2 & i > 1 \\
{[2^8]}^2 & i=1  \end{array} \right.$ & &\\
$i:(I-1) \rightarrow 1$ & &  & &  \\
\hline
\hline
Concat$_{v,i}$ &  &  & &\multirow{3}{*}{\includegraphics[width=.4cm]{T1Concat}}\\
$v:1 \rightarrow V$ & Concatenation &  $(U+1) \times 2^{(i+2)} \times [2^{(8-i)}]^2$ & &\\
$i=(I-2) \rightarrow 1$ & & &  & \\
\hline
\hline
Output$_v$ & & \multirow{2}{*}{ $[2^8]^2$} &&\\
$v=1 \rightarrow V$ & &  &&\\
\hline
\end{tabular}
\label{tab02}
\end{table*}

\begin{figure}
\centering
\includegraphics[width=.5\textwidth]{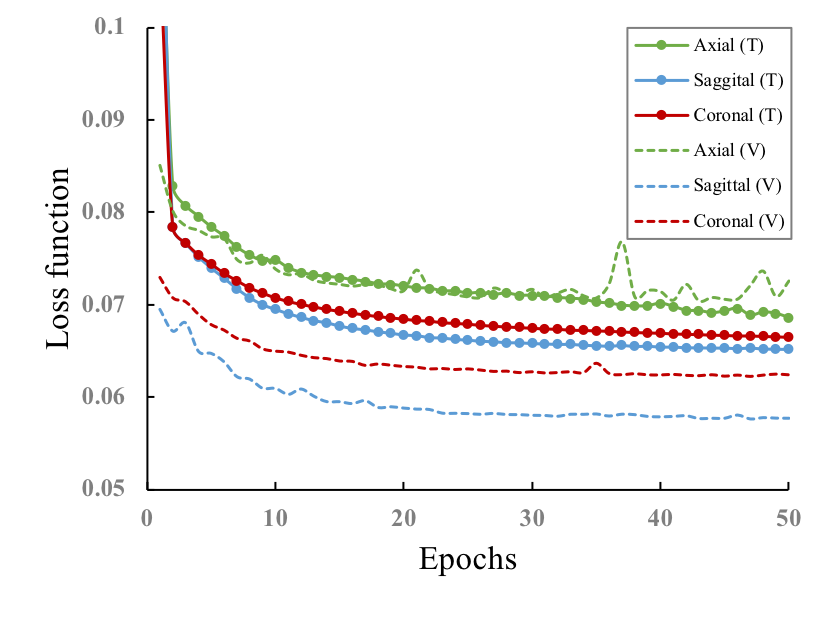}
\caption{Loss functions for training (T) and validation (V) for different slicing directions.}
\label{fig04}
\end{figure}


\subsection{Electromagnetic simulation}

The induced scalar potential $\psi$ in the volume head conductor is defined by the following equation:

\begin{equation}
\nabla.\sigma \nabla \psi = - \nabla.\sigma \frac{dA}{dt},
\label{eq4}
\end{equation}
where $A$ denotes the magnetic vector potential of the applied (external) magnetic field. The induced electric field is calculated as

\begin{equation}
E=-\nabla \psi - \frac{dA}{dt}.
\label{eq5}
\end{equation}

At intermediate frequencies, such as those in TMS applications, this formulation is valid if the electric and external magnetic fields are decoupled. Additionally, the conduction currents are at least one order of magnitude greater than the displacement currents, meaning only tissue conductivity is considered while permittivity is neglected~\cite{Dawson1996AC,Hirata2013PMB}. Equation~(\ref{eq4}) is solved numerically utilizing a scalar potential finite difference~\cite{Hirata2013PMB,Hirata2009RPD1} based on a multi-grid method with successive over-relaxation~\cite{Laakso2012PMB}. For comparison, we consider isotropic uniform tissue conductivity based on a fourth-order Cole-Cole model with a frequency of 10 kHz~\cite{Gabriel1996PMB}. Additionally, we consider other conductivity values that were used in previous studies as alternative values representing isotropic uniform conductivity. These values are listed in Table~\ref{tab01}~(B)~\cite{Aonuma2018neuroimage1}. Our computer simulation considered TMS using a figure-eight coil located above the scalp at position C3 (10-10 electroencephalogram international system) to target the hand motor area. The TMS coil was modeled using a thin-wire approximation with outer and inner diameters of 9.7 and 4.7 cm, respectively. The coil current was set equal to the maximum stimulation output of the TMS device.

\begin{figure}
\centering
\includegraphics[width=.45\textwidth]{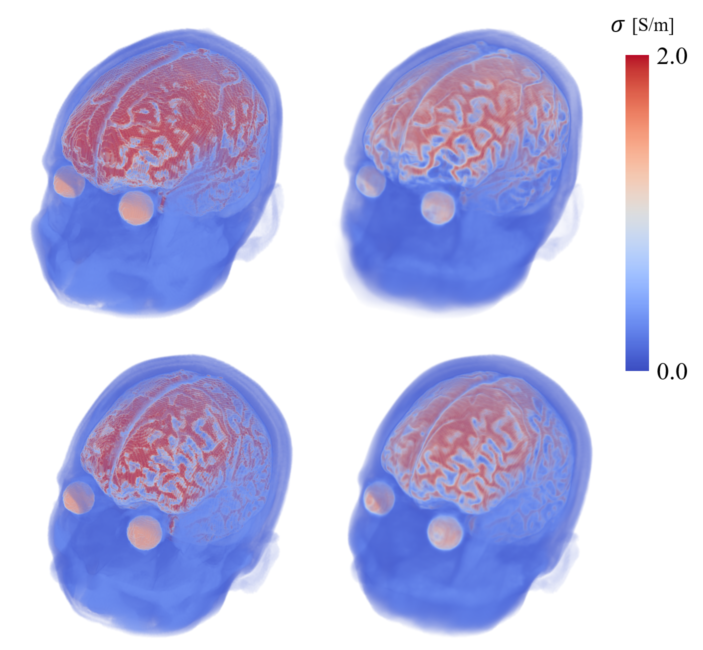}
\caption{The models on the left are volume conductors with uniform conductivity values computed using semi-automatic methods. The models on the right are volume conductors with non-uniform conductivity values computed using CondNet. The top models are (case01019) and the bottom models are (case01025). Two-dimensional slices with the corresponding MRI anatomies for (case01019) are presented in Fig.~\ref{fig06}.}
\label{fig05}
\end{figure}


\section{Results}


\subsection{Conductivity estimation}

\begin{figure*}
\centering
\includegraphics[width=\textwidth]{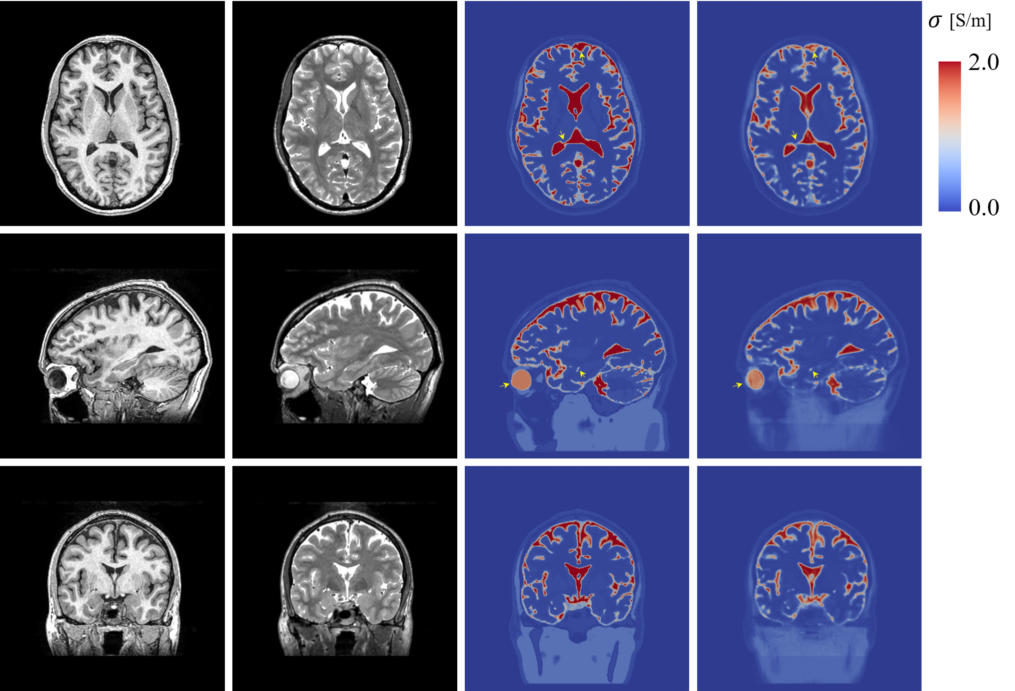}
\caption{From left to right: T1-weighted MRI, T2-weighted MRI, uniform conductivity maps (semi-automatic methods), and non-uniform conductivity maps (CondNet). From top to bottom: axial, sagittal, and coronal slices. Yellow arrows indicate regions in which the non-uniform conductivity values match the corresponding anatomy more closely. Regions corresponding to the neck were computed arbitrarily in CondNet images because the corresponding anatomy is unavailable.}
\label{fig06}
\end{figure*}

The proposed method was implemented using a workstation with four Intel (R) Xeon CPUs running at 3.60 GHz, 128 GB of memory, and three NIVIDIA GeForce GTX 1080 GPUs. The CondNet with $U=2$, $V=1$, and $I=6$ shown in Fig.~\ref{fig03} was implemented using Wolfram Mathematica\footnote{Wolfram Research, Inc., Champaign, IL, 2019} (R) ver. 12.0. The convolution kernels were set as follows: $R_{u,i}=3$ and $S_{v,i}=T_{v,i}=5$. We considered the conductivity values listed in Table~\ref{tab01}~(A) as true values for the training set. A set of 10 subjects was used to train three networks corresponding to different slicing directions. All slices were randomly shuffled and split with a ratio of 9:1 for training and validation, respectively. We considered cross-entropy loss function mapping, which was minimized using the ADAM algorithm~\cite{Kingma2014arXiv}. Training was conducted over 50 epochs with a batch size of four. A single training session was completed in less than 9 min for each slicing direction. The plot in Fig.~\ref{fig04} illustrates loss function convergence. One can see that validation in the axial direction performs differently compared to the other two directions. This behavior is expected because the ground-truth segmented models used for training the axial network include some features that were manually added to the neck region that have no corresponding anatomy in the MRI scans. The remaining eight subjects were used for evaluation. A volume render of two subjects is presented in Fig.~\ref{fig05} and a detailed example of one subject (case01019) is presented in Fig.~\ref{fig06}.   

From these results, one can see that CondNet is able to predict potential conductivity within a small fluctuation range around the uniform values. The distribution of conductivity values of the CondNet-generated volume conductors exhibits more realistic patterns with smooth transitions at tissue boundaries. Furthermore, there are some regions where the CondNet-generated volume conductors closely match the corresponding anatomy (Fig.~\ref{fig06}). For example, one can see that the conductivity values within the eye lens are estimated with lower values compared to the surrounding vitreous humor. It is worth noting that the segmented models used for training do not contain information regarding eye lens conductivity. However, when anatomical information is missing, the CondNet-generated conductivity values are somewhat arbitrary. This can be clearly observed in the lower neck regions of the conductivity maps presented in Fig.~\ref{fig06}, where the corresponding neck in the uniform volume conductor is more rigid. This region was manually added to the uniform volume conductor to compensate for the unavailability of anatomical data because these regions do not affect the electric field distributions in superior regions~ \cite{GomezTames2019JNE1}.

To demonstrate how the estimated non-uniform conductivity values are distributed at a small neighborhood scale, we compared the estimated conductivity values to standard uniform values. Seven regions of interest (ROIs) were selected from the central axial slice of (case01017) such that they contained homogeneous tissues based on the corresponding segmented model. ROIs were selected to represent samples of WM, GM, fat, bone (canc.), bone (cort.), muscle, and CSF. Both uniform and non-uniform values within limited neighborhoods are presented in Fig. 7. These results demonstrate excellent consistency when considering the variability of water contents in human tissues. For example, the water contents of WM and GM in adults are known to vary from 68\% to 77\% and 84\% to 86\%, respectively~\cite{Snyder1994Report}. Additionally, a wide range of conductivity values for different head tissues was presented in~\cite{McCann2019BT}.

In another experiment, the network architecture was altered such that $V=2$ (i.e., two conductivity maps were estimated simultaneously). This experiment aimed to estimate two different volume conductors based on different tissue conductivity values in a single shot. The network was trained using conductivity values obtained from the Cole-Cole model (A) and typical computational values (B) in Table~\ref{tab01}. The remaining CondNet parameters were similar to the previous experiment. The testing results and corresponding non-uniform models for (case01025) are presented in Fig.~\ref{fig08}. One can see that the CondNet-generated volume conductors exhibit a similar distribution of electrical conductivity values. In some regions, where differences can be observed, the CondNet results exhibit closer matches to the corresponding anatomical structures.

\begin{figure*}
\centering
\includegraphics[width=\textwidth]{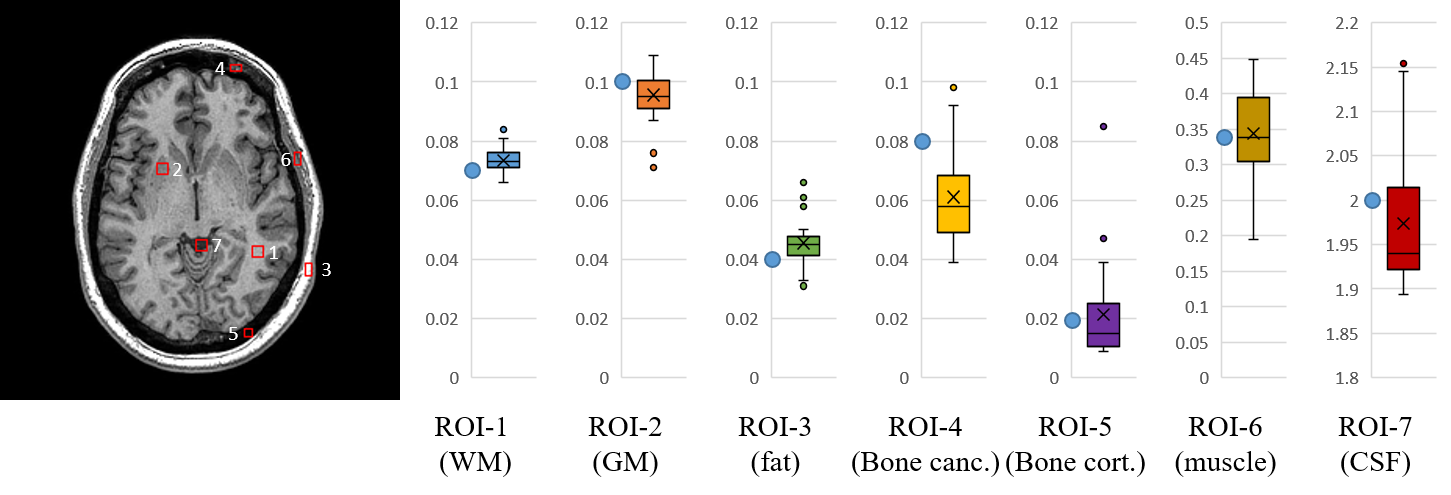}
\caption{ROI positions and labels in the central axial slice of (case01017) are presented on the left. The right side of the figure presents a boxplot of the conductivity values with uniform (blue circle over y axis) and non-uniform (box) volume conductors. The ROIs represent WM, GM, fat, bone (canc.), bone (cort.), muscle, and CSF in order.}
\label{fig07}
\end{figure*}


\subsection{TMS simulation}

A TMS-induced electric field was used to investigate the effects of the proposed approach. Electric fields were computed using the original uniform head model and network-generated conductivity models for eight subjects. The TMS coil was placed at position C3 to target the hand motor area. Figure~\ref{fig09} presents the electric fields for the cortical surface and a representative sagittal plane. Additionally, the differences between the electric fields in the hand motor area (target region) and sagittal plane are highlighted. The field distributions on the cortical surfaces are similar, but CSF content variations could affect spread and intensity because the network-generated model exhibits graded conductivity values between boundaries. For some subjects, the differences are more significant based on large discrepancies in the conductivity distributions, particularly for the target region. However, the differences are still within reasonable ranges considering the subject variability. 

To highlight the difference between electric field distributions, magnified regions of the sagittal slices of four subjects are presented in Fig. 10. The global distributions of the electric fields are consistent, but the models generated with non-uniform conductivity exhibit homogeneous patterns with fewer staircase artifacts. Global error (GE) is defined as follows:
\begin{equation}
GE=\frac{1}{\max_{i \in \Omega} (E_i, \hat{E}_i)}\times \frac{\sum_{i=1}^I  |E_i-\hat{E}_i|}{I} \times 100\%,
\end{equation}
where $E$ and $\hat{E}$ are the internal electric fields computed using uniform and non-uniform conductivity, respectively. The computed difference values are listed in Table~\ref{tab03} to quantify the electric field distributions in different regions. In the case of the hand motor area, the ROI was set to the hand knob region on the standard brain template centered on MNI (-42,-13,66) with a radius of 5 mm, which corresponds to the vicinity of the TMS hotspot for the abductor pollicis brevis muscle. The standard brain template ROI was projected onto each individual head model to obtain a corresponding ROI for each individual~\cite{Diekhoff2011HBM1}. The error is greatest in the hand motor area, where stronger electric fields are present and differences between models may be more significant based on the high conductivity contrast between tissues. In the brain, non-brain, and whole head regions, the global error is attenuated (conductivity contrast is smaller within the brain and other tissues).

\begin{figure*}
\centering
\includegraphics[width=\textwidth]{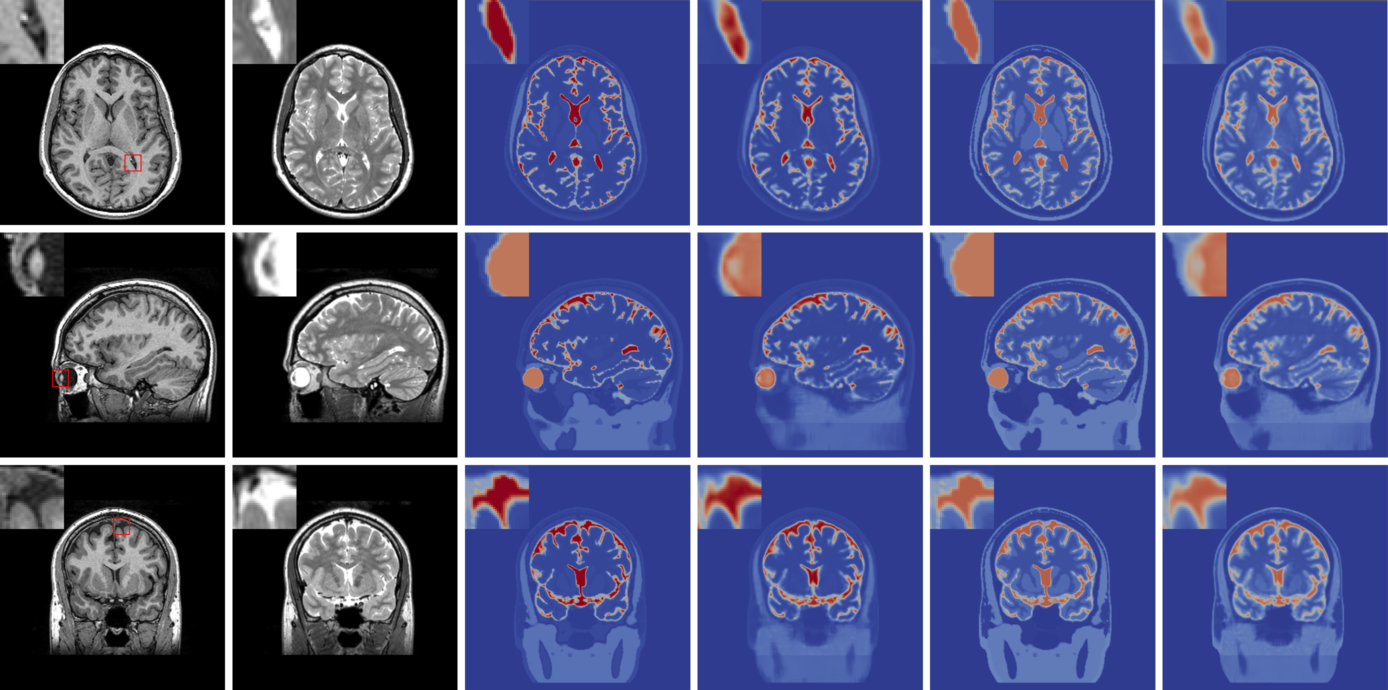}
\caption{Two sets of electrical conductivity maps corresponding to the values listed in Table~\ref{tab01} (A) and (B). Maps were generated using CondNet and compared to uniform conductivity maps. From left to right: T1-weighted MRI, T2-weighted MRI, uniform (A), non-uniform (A), uniform (B), and non-uniform (B) conductivity maps. From top to bottom: axial, sagittal, and coronal slices. The ROIs labeled in the leftmost column are magnified in the top-left corner of each image. It is clear from these ROIs that non-uniform conductivity maps provide a better representation of real anatomical structures.}
\label{fig08}
\end{figure*}

\begin{figure*}
\centering
\includegraphics[width=\textwidth]{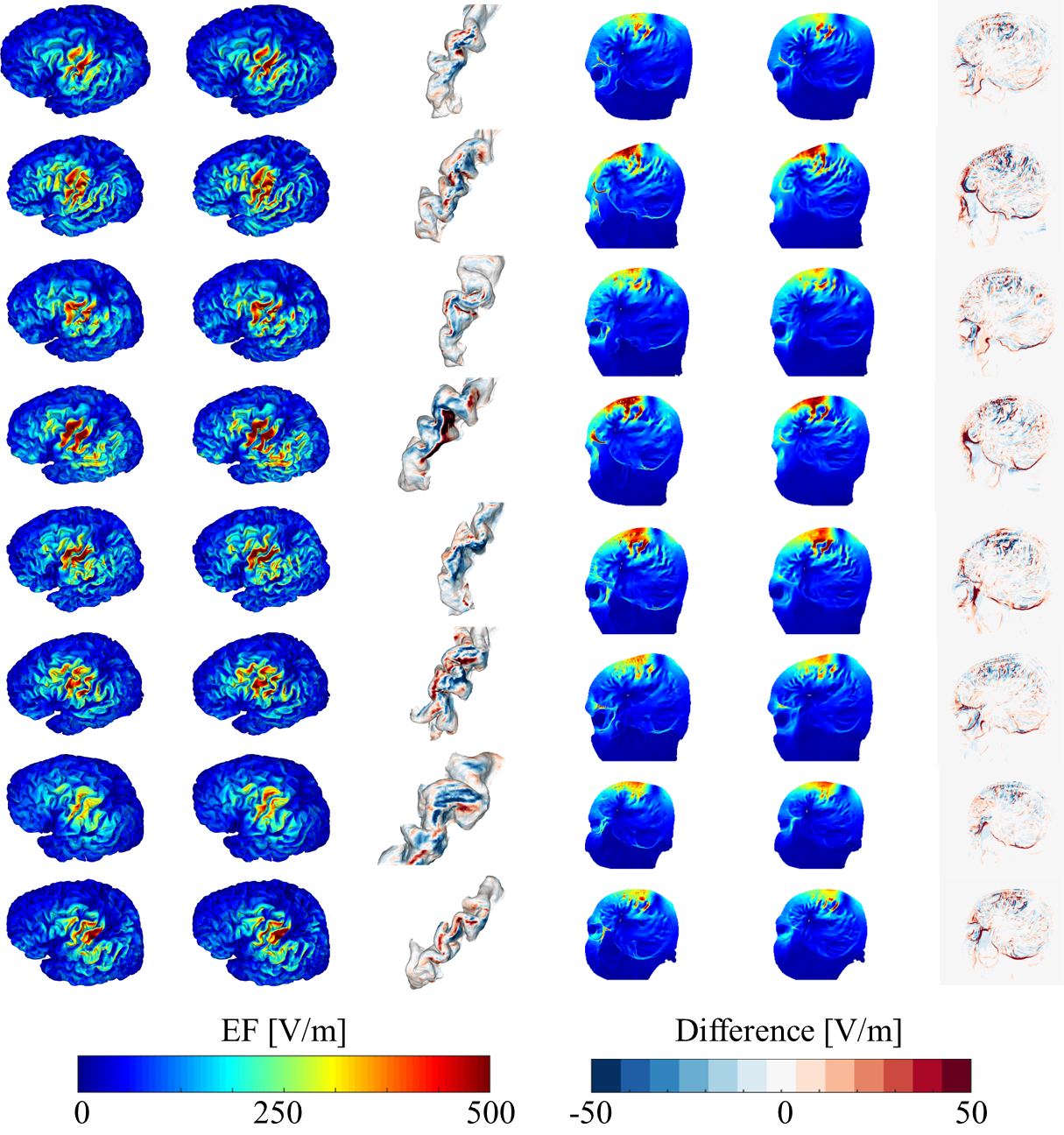}
\caption{Electric field distributions. The subjects listed in Table~\ref{tab03} are presented in order from top to bottom. The first two columns are the electric field maps for the brains using uniform ($E$) and non-uniform ($\hat{E}$) conductivity values, respectively, and third column represents the error within the hand motor area. The fourth and fifth columns are the electric field maps for the head (sagittal section) with uniform and non-uniform conductivity values. The sagittal section error ($E$ - $\hat{E}$) is presented in the rightmost column.}
\label{fig09}
\end{figure*}

\begin{table}
\centering
\footnotesize
\caption{GE of the electric fields in different groups of head tissues.}
\setlength{\tabcolsep}{3pt}
\begin{tabular}{ c l c c c c  }
\hline
\multirow{3}{*}{\#}&\multirow{3}{*}{\bf Subject}	& \multicolumn{4}{c}{\bf GE [\%] (mean$\pm$std)}   			\\
\cline{3-6}
&& \makecell{Motor Area \\(Hand knob)} & Brain & Non-Brain & Head\\
\hline
1&case01017        & 6.0$\pm$4.6  & 0.7$\pm$1.2 & 0.9$\pm$1.6 & 0.3$\pm$1.0 \\
2&case01019        & 5.5$\pm$4.2  & 0.8$\pm$1.4 & 1.1$\pm$2.3 & 0.4$\pm$1.2 \\
3&case01025        & 2.9$\pm$2.9  & 0.7$\pm$1.2 & 1.0$\pm$1.9 & 0.3$\pm$1.1 \\ 
4&case01028        & 6.0$\pm$4.2  & 0.8$\pm$1.4 & 0.8$\pm$1.7 & 0.3$\pm$1.0 \\
5&case01034        & 7.9$\pm$6.0  &	0.7$\pm$1.2 & 1.0$\pm$1.9 & 0.4$\pm$1.2 \\
6&case01039        & 7.7$\pm$5.4  &	0.7$\pm$1.3 & 0.9$\pm$1.8 & 0.3$\pm$1.0 \\
7&case01042        & 11.8$\pm$9.9 &	0.7$\pm$1.0 & 1.1$\pm$1.9 & 0.3$\pm$0.9 \\
8&case01045        & 6.0$\pm$4.5  &	0.6$\pm$1.1 & 0.9$\pm$1.8 & 0.3$\pm$1.0 \\
\hline
& average & 6.725 & 0.7125 & 0.9625 & 0.325 \\
\hline
\end{tabular}
\label{tab03}
\end{table}

\begin{figure*}
\centering
\includegraphics[width=\textwidth]{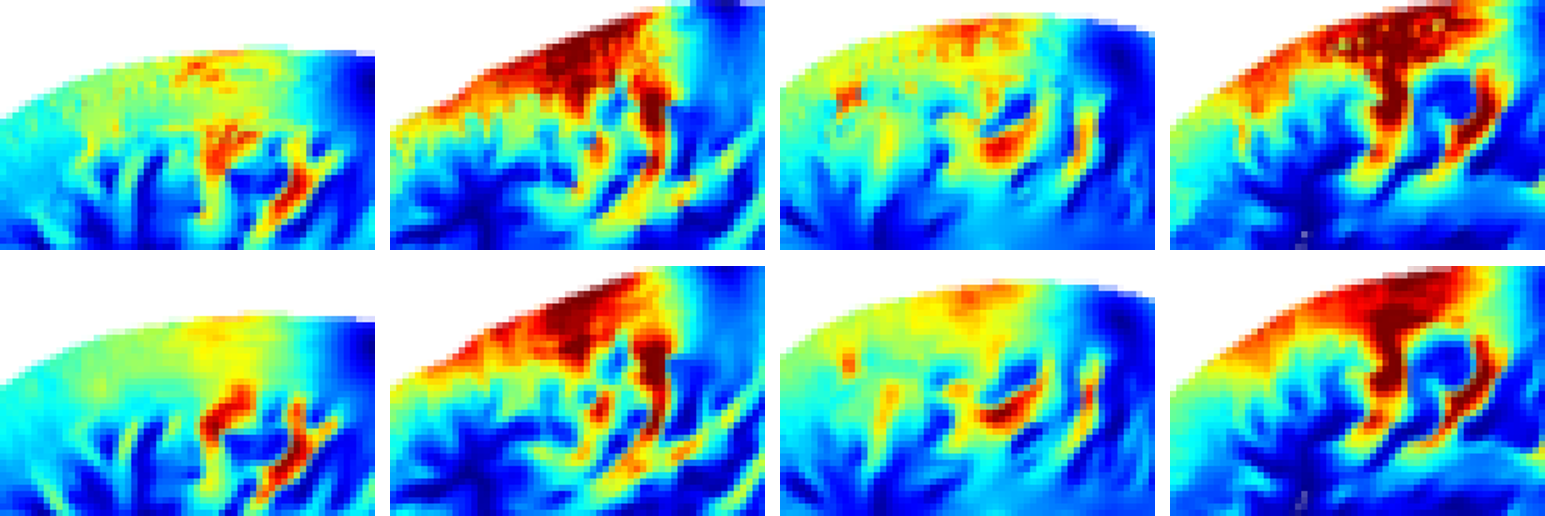}
\caption{Magnified cortical regions presented in Fig.~\ref{fig09}. From left to right are subjects case01017, 19, 25, and 28. The top and bottom rows are the electric field maps corresponding to uniform and non-uniform conductivity values, respectively. Highly consistent distributions of electric fields can be observed, but the maps generated using non-uniform conductivity values largely eliminate staircase artifacts.}
\label{fig10}
\end{figure*}


\section{Discussion}

Personalized TMS utilizing the current paradigm is a time-consuming process that is difficult to implement clinically. In a recent study, it was reported that TMS-induced electric fields have weak sensitivity to conductivity variations~\cite{Saturnino2019neuroimage}. Therefore, there is a frame in which non-uniform conductivity will produce comparable results. In this study, we developed a novel CNN architecture for the estimation of non-uniform electrical conductivity in human head models based on anatomical information extracted from T1- and T2-weighted MRI scans. The developed learning-based conductivity generator network can achieve high accuracy utilizing a limited training dataset containing only 10 subjects. This novel architecture provides several advantages. First, it does not require highly sophisticated imaging modalities and can estimate accurate conductivity values based on standard MRI scans. Second, CondNet provides conductivity maps that are highly consistent with anatomical structures. For example, the conductivity of eye lens that is not considered in training is estimated with value ranged between 0.31 and 0.55 (reference true value is 0.35~\cite{Gabriel1996PMB}). Third, employing CondNet can facilitate precision medicine in which a personalized head model can be generated in a short timeframe without time-consuming segmentation. Furthermore, it can be extended to estimate head models that are suitable for different stimulation scenarios in a single operation. Therefore, it can be used for brain stimulation planning for different clinical applications and the optimization of TMS dosing (e.g.,~\cite{GomezTames2018BS}).

Head models with non-uniform conductivity maps were evaluated based on TMS simulation of the hand motor area and compared to standard head models generated via segmentation with uniform conductivity. Generally, the patterns of electric field distributions generated by the two models were very consistent. However, upon closer observation, it was determined that staircase artifacts were present in the head models with uniform conductivity, whereas the CondNet models provided a more natural and uniform distribution of electric fields. Quantitative analysis demonstrated that a relatively small difference can be achieved, which can be referred to the staircase artifacts in models with uniform conductivity values.

The proposed architecture has a general form that can be extended in the future to include additional anatomical images with additional fine-tuning features, such as fiber orientations (DTI), blood vessels (venograms), and tumor activities (PET/SPECT). Additionally, it can be easily extended to generate models representing different electromagnetic properties, such as permittivity, as well as models that fit a wide range of frequencies. As shown in the results section, the proposed architecture is relatively fast in terms of both training (minutes) and testing (seconds), which should make it usable in clinical applications. It is expected that this approach will contribute to enhanced electromagnetic stimulation that can automatically estimate highly reliable conductivity values without time-consuming segmentation.


\section{Conclusion}

A novel CNN architecture was proposed for the automatic generation of human head models with non-uniform electric conductivity. The proposed CondNet was trained using uniform volume conductors and quickly estimated the non-uniform conductivity values of unknown subjects based on T1- and T2-weighted MRI scans. The proposed network has several merits in terms of enabling precision medicine applications for brain stimulation. First, intensive segmentation of multiple head tissues is not required, which reduces time and effort significantly compared to manual segmentation approaches. Second, as shown by the results presented in this paper, CondNet is able to estimate the conductivity values of anatomical structures that are not present in a training dataset. Third, CondNet has the ability to generate different volume conductors associated with multiple brain stimulation scenarios in a single operation. This should provide a flexible framework that can handle alternative stimulation conditions. The source code used in this study will be provided by the corresponding author upon receiving a reasonable request.

\bibliography{IEEEabrv,Refs1}
\end{document}